\pdfoutput=1
\documentclass[11pt]{article}

\PassOptionsToPackage{table}{xcolor}

\usepackage[final]{acl}

\usepackage[utf8]{inputenc}
\usepackage[T1]{fontenc}
\usepackage{endfloat}

\DeclareUnicodeCharacter{FF0C}{,}
\DeclareUnicodeCharacter{3001}{,}
\DeclareUnicodeCharacter{2013}{--}
\DeclareUnicodeCharacter{2014}{---}
\DeclareUnicodeCharacter{2212}{-}
\DeclareUnicodeCharacter{00A0}{~}
\DeclareUnicodeCharacter{2018}{'}
\DeclareUnicodeCharacter{2019}{'}
\DeclareUnicodeCharacter{201C}{"}
\DeclareUnicodeCharacter{201D}{"}
\DeclareUnicodeCharacter{FEFF}{}

\usepackage{graphicx}
\usepackage{booktabs,siunitx,threeparttable}
\usepackage{times}
\usepackage{latexsym}
\usepackage{amsmath,amssymb}
\usepackage{microtype}
\usepackage{inconsolata}
\usepackage{enumitem}

\title{Automated Item Neutralization for Non-Cognitive Scales: A Large Language Model Approach to Reducing Social-Desirability Bias}

\author{Sirui Wu \\
  University of British Columbia \\2125 Main Mall, \\Vancouver, BC V6T1Z4, Canada\\
  \texttt{sirui.wu@ubc.ca} \\\And
  Daijin Yang \\
    Northeastern University \\
  360 Huntington Ave,\\
  Boston, MA 02130, USA \\
  \texttt{yang.dai@northeastern.edu} \\}

\begin{document}
\maketitle
\begin{abstract}
This study evaluates item neutralization assisted by the large language model (LLM) to reduce social desirability bias in personality assessment. GPT-o3 was used to rewrite the International Personality Item Pool Big Five Measure (IPIP-BFM-50), and 203 participants completed either the original or neutralized form along with the Marlowe–Crowne Social Desirability Scale. The results showed a preserved reliability and a five-factor structure, with gains in conscientiousness and declines in Agreeableness and Openness. The correlations with social desirability decreased for several items, but inconsistently. Configural invariance held, though metric and scalar invariance failed. Findings support AI neutralization as a potential but imperfect bias-reduction method.
\end{abstract}

\section{Introduction}

Large language models have primarily been applied to generate cognitive test items and have shown strong performance. With proven powerful contextual understanding and generation abilities in multiple domains~\cite{fitria2023artificial,gpt4game,ullah2024challenges}, systems such as GPT-3~\cite{floridi2020gpt} have already produced acceptable multiple choice reading passages~\cite{shin2025exploratory}, chemistry and physics items~\cite{chan2025automatic}, and tasks that assess fluid reasoning and visual processing~\cite{ryoo2022development}. However, using LLMs for non-cognitive assessments (personality, attitudes, social-emotional skills) is still rare. These constructs are often abstract, value-laden, and context-dependent, which makes automatic item-writing challenging.  

Nonetheless, early research is beginning to explore this space.  \citet{li2024automatic} used GPT-4 to create short, scenario-based questions, named situational judgment items. These items ask people how they would respond in everyday situations, as a way of measuring the Big Five personality traits. In another example, \citet{pmlr-v264-xue25a} relied on GPT-3.5 to expand and translate a university-belongingness questionnaire, maintaining good reliability despite some noisy items. These findings suggest that LLMs can assist non-cognitive scale development, but their robustness and effectiveness remain unverified. Studies have shown that LLM outputs for complex social constructs, such as political or moral values, tend to be overly uniform~\cite{park2024diminished}. 

Most prior work has focused on generating new items from scratch, but refining existing validated items through targeted edits is an equally promising yet understudied approach.
As emphasized by The Standards for Educational and Psychological Testing~\cite{eignor2013standards}, adapting item wording --- whether for clarity, cultural context, or bias reduction --- can enhance accessibility and fairness while preserving construct validity. McCrae et al.~\cite{mccrae2005neo} demonstrate that systematic item refinement, like simplifying complex terms in the NEO-PI-3, improves readability and reliability without changing the test’s core structure. Studies also show that employing various refinement strategies, such as rephrasing and balancing item tone, can enhance validity while preserving construct discrimination~\cite{backstrom2014criterion}.

To research how LLM could serve as a precise editors, we adopted LLM to identify and decrease the social desirability bias.
Social-desirability bias is a tendency for a person to respond in a way that seems socially appealing, regardless of his or her true characteristic~\cite{sdb,Psychometrics,braun2001socially}. 
It can contaminate true levels of trait and comparison of individuals, especially on traits such as agreeableness, conscientiousness, and emotional stability~\cite{sdbbackstorm}, so curbing it is critical. 
It was chosen for our study not only because it is a common threat to non-cognitive tests, but also because traditional approaches to reducing social desirability, including forced choice~\cite{cao2019does}, balanced keying~\cite{gignac2013modeling,li2024mixed}, and manual ``neutralization'' of wording~\cite{backstrom2020properties,backstromsdpi}, can work but are labor intensive and may create unintended dimensions~\cite{zhang2025improving}.

Recent studies demonstrate that LLMs not only display social desirability response patterns similar to humans, but can also detect when they are being evaluated and shift their answers toward socially valued traits. This ability to recognize and reproduce bias suggests that LLMs could also be leveraged to diagnose and potentially mitigate social desirability effects in human surveys~\cite{lee2024exploring,salecha2024large}. For instance, ~\citet{dukanovic2025comparing} conducted a real-world hiring study. They required candidates completed both a standard multiple-choice personality questionnaire and a short conversation with an AI chatbot. The chatbot analyzed their written answers and generated personality scores, and they found chatbot-based scores were less influenced by social desirability than the traditional questionnaire scores. However, the chatbot scores were also less effective at predicting external outcomes such as education level or job role. Nevertheless, few studies have evaluated whether LLMs can rewrite test items to reduce their social desirability without compromising reliability and validity. 

To address this gap, we used prompt engineering to guide GPT-o3 in revising the IPIP-BFM-50~\cite{tao2009preliminary}, maintaining the test’s structure while reducing social desirability bias. The prompt integrated established debiasing strategies~\cite{kajonius2017cross,backstrom2014criterion} and incorporated role-playing~\cite{kong2023better}, chain-of-thought prompting~\cite{wei2022chain}, and transparency mechanisms~\cite{schneider2024explainable}. We evaluated the AI-neutralized items with participants against the original form, examining reliability, factor structure, and correlations with the Marlowe–Crowne Social Desirability Scale.

The results show that AI-based neutralization attenuated social desirability bias while preserving the Big Five structure within each form. Reliability was maintained in most domains, improved for Conscientiousness, but decreased for Agreeableness and Openness. Confirmatory factor analyses supported configural invariance, though full metric and scalar invariance across versions was not achieved. Correlations with social desirability weakened for many items, though effects were uneven across traits.

The discussion highlights both the promise and limitations of AI-assisted item editing. AI neutralization provides a viable tool for reducing response bias without altering trait constructs, but its uneven performance and lack of cross-form equivalence indicate the need for domain-specific fine-tuning, iterative refinement, and human-in-the-loop validations. Taken together, this work demonstrates the potential of large language models to contribute to fairer psychological assessment through targeted item rewriting.

\section {Methods}
\subsection{Instruments}
\subsubsection{The International Personality Item Pool Big Five Personality Scale (IPIP-BFM-50)}

We employed the IPIP-BFM-50 as the foundational measure of the Big Five personality traits, including 50 items~\cite{goldberg2006international,zheng2008reliability}. Each personality was measured by 10 items. This version of the IPIP-BFM-50 has been previously validated and shown to retain acceptable psychometric properties across multiple studies~\cite{tao2009preliminary}. Across multiple cultural validations, Cronbach’s alphas are generally high ($.80$ – $.90$) for Extraversion, Conscientiousness, Emotional Stability, and Openness, though Agreeableness is sometimes lower ($.65$ – $.70$)~\cite{goldberg2006international,ypofanti2015psychometric,zheng2008reliability}. Studies also observed validity evidence based on internal structure and relations to other scales. Factor analyses consistently replicate the expected five-factor structure, with strong invariance across gender and ethnic groups~\cite{constantinescu2016adaptation,buchanan2005implementing,ehrhart2008test}. Validity is shown through substantial correlations with other Big Five instruments including the NEO Five-Factor Inventory~\cite{gow2005goldberg}, the Ten Item Personality Inventory~\cite{ypofanti2015psychometric}, and the Eysenck Personality Questionnaire–Revised~\cite{gow2005goldberg}, often above $.60$. 

\subsubsection{IPIP-BFM-50 with AI-neutralization (IPIP-BFM-50-AI)}

To systematically reduce social desirability bias in personality assessments, we developed a tailored prompt for GPT-o3, producing the IPIP-BFM-50-AI. GPT-o3 was selected for its strong instruction following, long-context reasoning, and coherent, multi-step outputs~\cite{kim2025evaluating,ballon2025relationship,openai2025o3system}. Our design draws on Bäckström et al.’s manual rewriting strategies~\cite{backstrom2014criterion}, emphasizing reduced evaluative language, preserved behavioral meaning, and midpoint-oriented phrasing—methods shown to reduce item popularity while maintaining validity.

Beyond psychometric strategies, the prompt incorporates techniques to boost effectiveness and interpretability. It frames GPT-o3 as an expert psychometrician~\cite{kong2023better}, applies chain-of-thought prompting~\cite{wei2022chain} to structure reasoning, and enforces transparency through structured outputs with justifications, bias ratings, and fidelity checks~\cite{schneider2024explainable}. The full prompt and generated items are included in the appendix.

\subsubsection{The Marlowe‐Crowne Social Desirability Scale (MC-SDS) short form }

The SDS is a validated and widely used measure for assessing socially desirable responding. The Marlowe--Crowne Social Desirability Scale (MC-SDS) short forms, particularly the 13-item Reynolds version, exhibit acceptable internal consistency ($\alpha = .76$~\cite{reynolds1982development}) and very high correlations ($.80$--$.90$~\cite{ii1985cross}) with the full 33-item scale. This evidence supports their reliability and validity.

\subsection{Participants and Data Collection}

Participants were recruited online through public advertisements and social networks. Eligibility required age 18 or older, and consent to participate. After reading the study information page and providing informed consent, participants were randomly assigned to complete either the original or the AI neutralized version of the IPIP-BFM-50, creating a between-group design with two independent samples. The two forms used identical content domains but different wording where applicable for the AI-neutralized version. To prevent memory and sequence effects, item order was independently randomized within each version, and the version order was counterbalanced across participants. The response format used a 5-point radio-button scale for all items.
The Marlowe–Crowne Social Desirability Scale (short form) was administered after one of the two IPIP administrations. Demographic information (age, gender, education, occupation) was collected at the end to minimize priming. All participants were voluntary recruited by an online link, operated and delivered by a free online survey tool Wjx~\footnote{https://www.wjx.cn/}. We collected 203 response, 102 for Original version and 101 for Neutralized version. After excluding all cases with incomplete items, the sample size was finalized to be 200, each version with 100 responses.

\subsection{Evaluation Strategy and Hypothesis}

\subsubsection{Effectiveness of neutralization.}
\textbf{Item and scale level indicators.} We will compare item popularity (means, SDs) and scale means between original and AI-neutralized items to check that highly evaluative items show reduced extreme endorsement without loss of variability. 

\textbf{Desirability linkage.} Estimate the correlation of each domain with MC-SDS for the original and neutralized versions within persons. Test whether the neutralized version shows a smaller association with MC-SDS. 
\subsubsection{Validity evidence following the Standards.}
\textbf{Internal structure and reliability.} For each version, test unidimensionality within each domain via CFA or IRT dimensionality checks, then test cross-version invariance (configural, metric, scalar) and report reliability (Cronbach alpha).

\textbf{Relations to other variables.} As discriminant evidence, verify that neutralized scales show weaker correlations with social desirability than originals, while preserving expected convergent patterns with established Big Five constructs.

\subsubsection{Hypothesis}
Results from all analysis above can be used to check the following hypothesis:
\begin{enumerate}
    \item \textbf{H1 - reliability}: Neutralized domains will demonstrate acceptable reliability that is comparable to originals.
    \item \textbf{H2 - structure}: Each domain will show a single intended factor per version and acceptable cross-version invariance indices.
    \item \textbf{H3 - relations}: Neutralized domains will maintain expected convergent patterns with Big Five constructs while showing reduced linkage to social desirability.
\end{enumerate}

\section{Results}
Two balanced groups completed the original and AI-neutralized versions (\textit{n} = 100 each). As shown in Table \ref{tab:socio-demo}, most participants were between 26 and 40, and there were also respondents aged from 41 to 50, as well as a small 60+ group. Gender distributions were comparable across versions, with roughly equal numbers of men and women. The groups appear demographically similar, supporting a fair comparison of psychometric results between original and neutralized items.

\begin{table}[ht]
\centering
\caption{Socio-demographics Variable}
\label{tab:socio-demo}
\footnotesize 
\begin{tabular}{lcc}
\hline
\textbf{Age}      & \textbf{Original count} & \textbf{Neutralized count} \\ \hline
18--25            & 9  & 13 \\
26--30            & 17 & 24 \\
31--35            & 59 & 52 \\
36--40            & 8  & 6  \\
41--50            & 6  & 3  \\
Over 60           & 1  & 2  \\ \hline
\textbf{Gender}   &     &     \\ \hline
Male              & 49 & 46 \\
Female            & 51 & 54 \\ \hline
\end{tabular}
\end{table}

\subsection{H1: Reliability}
Reliability was largely preserved after neutralization. As shown in Table \ref{tab:reliability}, extraversion and Neuroticism remained high in both versions. Conscientiousness improved in the neutralized form. Agreeableness and Openness decreased, with Agreeableness dropping to the mid $.50s$ – $.60s$. Overall, alpha and omega were acceptable for most domains, indicating that neutralization did not broadly undermine internal consistency, though Agreeableness warrants caution. These findings support H1 with noted caveats. 
\begin{table}[ht]
\centering
\caption{Reliability for All Subscales}
\label{tab:reliability}

\small 
\begin{tabular}{lcccc}
\hline
 & \multicolumn{2}{c}{\textbf{Original}} & \multicolumn{2}{c}{\textbf{Neutralized}} \\ \cline{2-5}
\textbf{Subscale} & \textbf{Alpha} & \textbf{Omega} & \textbf{Alpha} & \textbf{Omega} \\ \hline
Extraversion        & 0.90 & 0.91 & 0.87 & 0.89 \\
Agreeableness       & 0.67 & 0.71 & 0.59 & 0.63 \\
Conscientiousness   & 0.73 & 0.77 & 0.79 & 0.81 \\
Neuroticism         & 0.91 & 0.91 & 0.94 & 0.94 \\
Openness            & 0.78 & 0.78 & 0.66 & 0.71 \\ \hline
\end{tabular}
\end{table}

\subsection{H2: The Validity Evidence from Internal Structure}

Single-group confirmatory factor analyses (CFAs) supported the intended five-factor structure for each version. As shown in Table \ref{tab:cfa-fit}, model fit was acceptable for the original version (\( \text{CFI} \approx .97, \ \text{TLI} \approx .97, \ \text{RMSEA} \approx .06 \)) and marginally weaker for the neutralized version (\( \text{CFI} \approx .97, \ \text{TLI} \approx .96, \ \text{RMSEA} \approx .08 \)). Both versions retain the five-factor structure, but the higher RMSEA in the neutralized form points to a few items needing targeted wording revision.

Multi-group tests showed that configural form held, but metric and scalar constraints produced significant misfit with elevated RMSEA, indicating a lack of full cross-version equivalence. Thus, Hypothesis~2 is partially supported: the structure replicates within versions, but strict invariance across versions was not achieved. Configural invariance was supported, indicating that the neutralized and original versions share the same five-factor pattern and item-to-factor assignments. This shows that neutralization preserved the construct blueprint. However, subsequent metric and scalar constraints did not hold, which implies differences in loadings and intercepts across forms. Scores can be interpreted within each form using the same domain structure, but cross-form comparisons of factor means should be deferred until partial invariance or alignment is applied.

\begin{table*}[ht]
\centering
\caption{Confirmatory Factor Analysis Model Fit on the Big Five Personality Model}
\label{tab:cfa-fit}
\small 
\renewcommand{\arraystretch}{1.3}
\begin{tabular}{lccccccc}
\multicolumn{8}{l}{\textbf{Single-group CFA fit}} \\ \hline
\textbf{Group} & \(\boldsymbol{\chi^{2}}\) (scaled) & \textbf{df} & \textbf{p (scaled)} & \textbf{CFI} & \textbf{TLI} & \textbf{RMSEA [90\% CI]} & \textbf{SRMR} \\ \hline
Original      & 1284.957 & 1165 & $<$0.001 & 0.972 & 0.971 & 0.060 [0.033, 0.079] & 0.093 \\
Neutralized   & 1336.244 & 1165 & $<$0.001 & 0.965 & 0.963 & 0.078 [0.055, 0.097] & 0.101 \\ \hline
\\[-1.2ex]
\multicolumn{8}{l}{\textbf{Multi-group invariance}} \\ \hline
\textbf{Model} & \textbf{Df} & \textbf{AIC} & \textbf{BIC} & \(\boldsymbol{\chi^{2}}\) & \(\boldsymbol{\Delta df}\) & \(\boldsymbol{\Delta \chi^{2}}\) & \textbf{p} \\ \hline
Configural    & 2330 & 25254 & 26309 & 4576.3 &  ---  &   ---   &   ---   \\
Metric        & 2375 & 25361 & 26268 & 4773.8 &  45   & 197.51 & $<$0.001 \\
Scalar        & 2420 & 25624 & 26383 & 5127.0 &  45   & 353.16 & $<$0.001 \\ \hline
\end{tabular}
\end{table*}

\subsection{H3: The Validity Evidence from Relations to Other Variables}
As shown in Table \ref{tab:diff-corr}, the results highlight differences in correlations between individual items across the five dimensions and the SDR score. It is expected to observe a decrease in difference for absolute value of correlation (no matter a positive or negative), indicating a decrease of influence by SDR. However, we can observe correlations are increase for some items. We conduct the Steiger’s Z test to check whether the change in correlation significant, and 6 items indicate a significant change. Among them, five are decrease and one increase. 

The neutralized items demonstrated reduced associations in several cases, supporting the intended effect. However, the presence of increases underscores uneven performance across content. Overall, H3 is partially supported: the tool attenuates social desirability bias for many items, but not consistently across the full instrument. 

Table \ref{tab:item-sdr} further showed details about what items was assessed to have significant change in correlations with SDR after neutralization. The changes align with specific linguistic mechanisms. For Extraversion, neutralized phrasings replace overt status claims with modest, observable behaviors or internal states. This lowers self-presentational stakes and reduces the incentive to answer in a socially approved way. For Openness, edits remove prestige cues (for example, “rich vocabulary”) and normalize difficulty with abstract content. Endorsing these becomes less face-threatening, so links to desirability weaken. The Agreeableness increase arises from hedged, evaluative wording (“others might find rude” and “sometimes”). This introduces norm salience and plausible deniability, inviting impression management more than the blunt behavior label “insult people.” In short, SDR decreases when wording is concrete, behavioral, and low in status or virtue signals; SDR increases when wording invokes social judgment, hedges frequency, or allows reframing of intent.

To sum up, AI neutralization works, but not uniformly. It maintains reliability in most domains, preserves the factor structure within forms, and reduces desirability in several areas. The costs are local and fixable: a handful of items drive non-invariance and dips in Agreeableness and Openness. Treat scores as within-form for now, apply partial invariance or alignment for cross-form comparisons, and revise the flagged items to restore behavioral precision while keeping neutral tone.

\begin{table*}[ht]
\centering
\caption{Difference in the Correlations with the SDR}
\label{tab:diff-corr}
\small
\renewcommand{\arraystretch}{1.3} 
\begin{tabular}{lccccc}
\hline
 & \textbf{Extraversion} & \textbf{Agreeableness} & \textbf{Conscientiousness} & \textbf{Neuroticism} & \textbf{Openness} \\ \hline
\textbf{delta} & -0.14 & 0.06 & -0.11 & 0.12 & 0.00 \\
\textbf{p}     & 0.26  & 0.66 & 0.43  & 0.34 & 0.98 \\ \hline
\textbf{delta} & -0.09 & -0.03 & -0.01 & 0.11 & \cellcolor{gray!30}-0.29 \\
\textbf{p}     & 0.47  & 0.84  & 0.96  & 0.41 & \cellcolor{gray!30}0.03* \\ \hline
\textbf{delta} & 0.16  & \cellcolor{gray!30}0.28 & 0.07  & 0.04 & \cellcolor{gray!30}-0.29 \\
\textbf{p}     & 0.19  & \cellcolor{gray!30}0.03* & 0.58  & 0.77 & \cellcolor{gray!30}0.04* \\ \hline
\textbf{delta} & \cellcolor{gray!30}-0.45 & 0.10  & 0.03  & 0.16 & -0.03 \\
\textbf{p}     & \cellcolor{gray!30}$<$0.001 & 0.49 & 0.83 & 0.20 & 0.81 \\ \hline
\textbf{delta} & -0.12 & -0.12 & 0.10  & -0.13 & 0.21 \\
\textbf{p}     & 0.37  & 0.36  & 0.41  & 0.29  & 0.14 \\ \hline
\textbf{delta} & -0.13 & 0.00  & -0.05 & 0.14  & -0.02 \\
\textbf{p}     & 0.29  & 0.98  & 0.71  & 0.24  & 0.86 \\ \hline
\textbf{delta} & -0.14 & -0.11 & -0.05 & 0.06  & 0.06 \\
\textbf{p}     & 0.28  & 0.37  & 0.72  & 0.60  & 0.62 \\ \hline
\textbf{delta} & -0.24 & -0.09 & 0.09  & 0.18  & -0.07 \\
\textbf{p}     & \cellcolor{gray!30}0.05* & 0.53  & 0.45  & 0.16  & 0.64 \\ \hline
\textbf{delta} & -0.07 & 0.08  & 0.18  & 0.05  & -0.26 \\
\textbf{p}     & 0.58  & 0.58  & 0.20  & 0.68  & 0.06 \\ \hline
\textbf{delta} & 0.10  & 0.01  & 0.00  & 0.04  & 0.14 \\
\textbf{p}     & 0.44  & 0.94  & 0.98  & 0.73  & 0.28 \\ \hline
\end{tabular}
\end{table*}

\begin{table*}[ht]
\centering
\caption{Original and neutralized items with SDR correlation changes}
\label{tab:item-sdr}
\small
\renewcommand{\arraystretch}{1.25}
\begin{tabular}{p{0.11\linewidth} p{0.39\linewidth} p{0.13\linewidth} p{0.10\linewidth} p{0.15\linewidth}}
\hline
\textbf{Version} & \textbf{Items} & \textbf{Personality} & \textbf{Direction} & \textbf{Correlation with SDR after neutralization} \\
\hline
Original    & Don't mind being the center of attention.         & Extraversion  & Positive &   \\
Neutralized & Feel fine when attention is on me.                & Extraversion  & Positive & Decrease \\
\hline
Original    & Am the life of the party.                         & Extraversion  & Positive &   \\
Neutralized & Often take an active role in group conversations. & Extraversion  & Positive & Decrease \\
\hline
    Original    & Insult people.                                & Agreeableness & Negative &   \\
Neutralized & Sometimes say things that others might find rude. & Agreeableness & Negative & Increase \\
\hline
Original    & Have a rich vocabulary.                           & Openness      & Positive &   \\
Neutralized & Know and use a variety of words.                  & Openness      & Positive & Decrease \\
\hline
Original    & Have difficulty understanding abstract ideas.     & Openness      & Negative &   \\
Neutralized & Find abstract ideas challenging.                  & Openness      & Negative & Decrease \\
\hline
\end{tabular}
\end{table*}

\section{Discussion}

The findings indicate that AI-based neutralization can reduce socially desirable responding while preserving the intended construct structure of a Big Five inventory. Single-group CFAs recovered the five-domain pattern in both versions, which suggests that the core representation of the traits remained intact after neutralization. Multi-group analyses supported configural invariance but not metric or scalar invariance, which implies that some item–factor relations and intercepts changed across versions. Reliability remained acceptable for most domains, improved for Conscientiousness, and declined for Agreeableness and Openness. Associations with a social desirability criterion decreased for several items, with notable exceptions in Agreeableness. Together, the results support AI neutralization as a viable wording intervention that targets response bias without altering trait identity.

\subsection{Construct representation and measurement comparability}

The preserved five-factor structure indicates that neutralization did not shift the meaning of the constructs, which aligns with evidence that the Big Five structure is robust across formats and raters \cite{mccrae1987validation}. The lack of metric and scalar invariance signals that item functioning changed across versions, so cross-form comparisons of means should not be made without partial invariance or alignment solutions \cite{byrne1989testing, putnick2016measurement}. Within each form, factors can be interpreted in the usual way. Across forms, unit and intercept differences should be addressed before comparing group or condition means.

\subsection{Domain-specific reliability shifts}
Conscientiousness reliability increased in the neutralized form, which is consistent with the idea that removing evaluative phrasing can sharpen behavioral focus and raise inter-item coherence. Declines in Agreeableness and Openness suggest that some edits broadened meanings or removed construct-diagnostic cues that previously fostered homogeneity. This pattern is compatible with prior work showing that evaluative wording can inflate internal consistency by cueing a general “goodness” factor, and that neutralizing language can reduce that inflation while leaving substantive variance intact \cite{backstrom2014criterion, backstromsdpi}.

\subsection{Why correlation with SDR changed}
Reductions in correlation with social desirability appear, when wording shifts from status or virtue claims to concrete behaviors or internal states. This likely weakens impression management, which is one facet of socially desirable responding \cite{paulhus1991enhancement}. Increases were observed when neutralized items introduced hedges or explicit social judgment cues, which can heighten norm salience and invite self-presentation. These mechanisms align with research on common method bias and evaluative content as drivers of spurious covariance and inflated correlations \cite{podsakoff2003common, backstromsdpi}.

\subsection{Implications for AI-assisted item editing}
The results indicate the potential of AI-assisted item editing. Recent research has shown that LLMs themselves exhibit human-like social desirability biases when responding to personality questionnaires, which implies that they are sensitive to the evaluative cues embedded in item wording and may therefore be leveraged to identify and mitigate such bias~\cite{chan2025automatic}. This capacity provides a foundation for the observed reduction in socially desirable responding when items are neutralized with AI support.

However, as the results suggest, a one-time output from a single prompt may not achieve the ideal output. Studies across multiple domains have found that single-shot generation often produces variable quality and is less reliable for tasks requiring precision, nuance, or consistency~\cite{patel2023limits,sahoo2024systematic}. The variability is partly due to the probabilistic nature of LLMs and the difficulty of capturing subtle linguistic properties in a single attempt. Research on prompting and iterative generation shows that multiple candidates and refinement loops generally outperform one-shot outputs, which supports the interpretation that item editing requires more than a single pass~\cite{iteration1,iteration2}.

Besides using single prompts, other techniques for enhancing large language model behavior are suggested. For the model itself, domain-specific fine-tuning has been shown to substantially improve performance even when only a small amount of high-quality training data is available~\cite{jeong2024fine,satterfield2024fine}. In this context, including pairs of successfully human-edited and neutralized items could increase the model’s ability to generate valid revisions. However, such data are difficult to obtain, and constructing this type of dataset is therefore an important future direction.

To add control to the system, multiple agents can be combined to provide feedback and review of generated items. One approach is to use another large language model as a reviewer, which can rate and critique generated items. Generate–feedback loops of this kind have proven effective in other domains, such as reasoning and dialogue, by reinforcing higher quality outputs through self-critique and refinement~\cite{li2024llms,madaan2023self}. Beyond automated feedback, incorporating humans in the loop transforms item generation into an iterative process. In such a cycle, participants test the items, results are analyzed, and the items are further refined based on psychometric evidence. This practice reflects established best practices in test development~\cite{eignor2013standards}, where iterative pilot testing and expert review are essential to ensure reliability and validity. Yet, the human-LLM collaboration still remains unexplored in the item editing field.

In summary, the results highlight the potential of AI-assisted item editing but also point to current limitations when relying on single-prompt outputs. Future development will benefit from domain-specific fine-tuning, multi-agent or human-in-the-loop feedback mechanisms, and iterative refinement processes that mirror traditional psychometric standards. Together, these strategies can convert AI neutralization into a reproducible pipeline that reduces bias while maintaining the measurement of intended psychological constructs.

\subsection{Limitations}

The study used a single language, a single instrument, and a between-groups design in a low-stakes context. Social desirability effects can be stronger under incentives to self-present, which limits generalizability to high-stakes settings. All measures were self-report and collected in one session, which raises the possibility of common method variance despite anonymity instructions. The analyses focused on internal structure, reliability, and associations with a bias criterion, so criterion-related validity with external outcomes remains untested for the neutralized form.

Future work should test neutralized items in high-stakes contexts, use within-person designs to estimate per-respondent reductions in bias, and include informant or behavioral criteria to address common method concerns. Partial invariance searches or alignment should be applied to enable cross-form comparisons, and results should document the number and type of freed parameters. The AI pipeline should be benchmarked across models and prompts, with a reusable library of prompt patterns and failure cases by domain. Replication across languages and populations, test–retest studies, and evaluation of predictive validity will clarify whether bias reduction is achieved without loss of criterion-related information.

\section{Conclusion}

AI-based neutralization reduced social desirability bias while preserving the Big Five construct structure. Reliability shifts varied across domains, improving for Conscientiousness but declining for Agreeableness and Openness, reflecting the influence of evaluative language on internal consistency. Configural invariance was supported, but metric and scalar invariance were not, indicating that cross-form comparisons require partial invariance or alignment methods. The discussion highlights that AI-assisted item editing is promising but uneven, and future development should emphasize domain-specific fine-tuning, iterative refinement, and human-in-the-loop validation to ensure stable and valid measurement.

\appendix

\section{Appendix}
\label{sec:appendix}
\subsection*{The Prompt for Neutralizing Self-Report Items}

You are an expert psychometrician. Your goal is to reword self‑report survey items so they measure the intended vocational interest while minimizing social desirability bias.

Social desirability bias is a type of response bias in research where participants tend to answer questions in a way that they believe will be viewed favorably by others, rather than providing completely honest or truthful responses.

Follow these rules:

\begin{itemize}
  \item Evaluate each item's social desirability bias. Give each item a score within -5 to 5 where 0 represents the lowest social desirability bias, 5 represents positive social desirability bias (people want to choose the item because they think the item is favorable), and -5 represents negative social desirability bias (people do not want to choose the item because they think the item is unfavorable). Keep the item unchanged if its social desirability bias score is in the zone from -1 to 1.

  \item Think step‑by‑step --- identify value‑laden terms, propose alternatives, and self‑check that the new wording still reflects the original behaviour, and that the new wording reduces the social desirability bias --- but \textbf{do not reveal your reasoning}.
  
  \item Remove or soften status-, value-, or social desirability‑laden words.

  \item Construct an item that you would find less desirable yourself.
  
  \item If the adjective is evaluatively positive, use a less evaluative one, or rephrase in a way that makes the adjective less evaluative.
  
  \item Do not change an item from positive to negative (direction).
  
  \item Think of whether the item is reversed or not.
  
 \item Preserve each item’s core behavioural meaning.
 
   \item Pay attention to the dimension of each statement. Do \textbf{NOT} change the dimension of each statement.
  
  \item Explain your change in natural language for each statement, and give your change a score to indicate its new social desirability bias.

\end{itemize}

\noindent\textbf{Output format:}

\noindent Please output the results in a 5-column table titled \textbf{Neutralized Items}, with the following headers:

\begin{center}
\textbf{Original \quad | \quad SD Score \quad | \quad Neutralized \quad | \quad Reason \quad | \quad SD Score}
\end{center}

\noindent Each statement is tagged with a dimension based on the Big Five personality traits. Use the following codes:

\begin{itemize}
  \item A: Agreeableness
  \item C: Conscientiousness
  \item N: Neuroticism
  \item O: Openness to Experience
  \item E: Extraversion
\end{itemize}

\noindent The sign "+" or "–" indicates whether the item is positively or negatively phrased within that dimension.


\begin{thebibliography}{56}
\providecommand{\natexlab}[1]{#1}

\bibitem[{B\"{a}ckstr\"{o}m and Bj\"{o}rklund(2014)}]{sdbbackstorm}
Martin B\"{a}ckstr\"{o}m and Fredrik Bj\"{o}rklund. 2014.
\newblock \href {https://doi.org/10.1027/1614-0001/a000138} {Social desirability in personality inventories}.
\newblock \emph{Journal of Individual Differences}, 35(3):144--157.

\bibitem[{B{\"a}ckstr{\"o}m and Bj{\"o}rklund(2020)}]{backstrom2020properties}
Martin B{\"a}ckstr{\"o}m and Fredrik Bj{\"o}rklund. 2020.
\newblock The properties and utility of less evaluative personality scales: Reduction of social desirability; increase of construct and discriminant validity.
\newblock \emph{Frontiers in psychology}, 11:560271.

\bibitem[{B{\"a}ckstr{\"o}m et~al.(2014)B{\"a}ckstr{\"o}m, Bj{\"o}rklund, and Larsson}]{backstrom2014criterion}
Martin B{\"a}ckstr{\"o}m, Fredrik Bj{\"o}rklund, and Magnus~R Larsson. 2014.
\newblock Criterion validity is maintained when items are evaluatively neutralized: Evidence from a full--scale five--factor model inventory.
\newblock \emph{European Journal of Personality}, 28(6):620--633.

\bibitem[{Ballon et~al.(2025)Ballon, Algaba, and Ginis}]{ballon2025relationship}
Marthe Ballon, Andres Algaba, and Vincent Ginis. 2025.
\newblock The relationship between reasoning and performance in large language models--o3 (mini) thinks harder, not longer.
\newblock \emph{arXiv preprint arXiv:2502.15631}.

\bibitem[{Braun et~al.(2001)Braun, Jackson, and Wiley}]{braun2001socially}
Henry~I Braun, Douglas~N Jackson, and David~E Wiley. 2001.
\newblock Socially desirable responding: The evolution of a construct.
\newblock In \emph{The role of constructs in psychological and educational measurement}, pages 61--84. Routledge.

\bibitem[{Buchanan et~al.(2005)Buchanan, Johnson, and Goldberg}]{buchanan2005implementing}
Tom Buchanan, John~A Johnson, and Lewis~R Goldberg. 2005.
\newblock Implementing a five-factor personality inventory for use on the internet.
\newblock \emph{European Journal of Psychological Assessment}, 21(2):115--127.

\bibitem[{Byrne et~al.(1989)Byrne, Shavelson, and Muth{\'e}n}]{byrne1989testing}
Barbara~M Byrne, Richard~J Shavelson, and Bengt Muth{\'e}n. 1989.
\newblock Testing for the equivalence of factor covariance and mean structures: the issue of partial measurement invariance.
\newblock \emph{Psychological bulletin}, 105(3):456.

\bibitem[{Bäckström and Björklund(2013)}]{backstromsdpi}
Martin Bäckström and Fredrik Björklund. 2013.
\newblock \href {https://doi.org/10.1111/sjop.12015} {Social desirability in personality inventories: Symptoms, diagnosis and prescribed cure}.
\newblock \emph{Scandinavian Journal of Psychology}, 54(2):152--159.

\bibitem[{Cao and Drasgow(2019)}]{cao2019does}
Mengyang Cao and Fritz Drasgow. 2019.
\newblock Does forcing reduce faking? a meta-analytic review of forced-choice personality measures in high-stakes situations.
\newblock \emph{Journal of Applied Psychology}, 104(11):1347.

\bibitem[{Chan et~al.(2025)Chan, Ali, Park, Sham, Tan, Chong, Qian, and Sze}]{chan2025automatic}
Kuang~Wen Chan, Farhan Ali, Joonhyeong Park, Kah Shen~Brandon Sham, Erdalyn Yeh~Thong Tan, Francis Woon~Chien Chong, Kun Qian, and Guan~Kheng Sze. 2025.
\newblock Automatic item generation in various {STEM} subjects using large language model prompting.
\newblock \emph{Computers and Education: Artificial Intelligence}, 8:100344.

\bibitem[{Cheng et~al.(2024)Cheng, Chen, Huang, Xing, Xu, and Lu}]{iteration1}
Yu~Cheng, Jieshan Chen, Qing Huang, Zhenchang Xing, Xiwei Xu, and Qinghua Lu. 2024.
\newblock \href {https://doi.org/10.1145/3638247} {Prompt sapper: A {LLM}-empowered production tool for building {AI} chains}.
\newblock \emph{ACM Trans. Softw. Eng. Methodol.}, 33(5).

\bibitem[{Constantinescu and Constantinescu(2016)}]{constantinescu2016adaptation}
PM~Constantinescu and I~Constantinescu. 2016.
\newblock The adaptation of the big-five {IPIP}-50 questionnaire in romania revisited.
\newblock \emph{Bulletin of the Transilvania University of Bra{\c{s}}ov. Series VII: Social Sciences• Law}, pages 129--138.

\bibitem[{Dukanovic and Krpan(2025)}]{dukanovic2025comparing}
Danilo Dukanovic and Dario Krpan. 2025.
\newblock Comparing chatbots to psychometric tests in hiring: reduced social desirability bias, but lower predictive validity.
\newblock \emph{Frontiers in Psychology}, 16:1564979.

\bibitem[{Ehrhart et~al.(2008)Ehrhart, Roesch, Ehrhart, and Kilian}]{ehrhart2008test}
Karen~Holcombe Ehrhart, Scott~C Roesch, Mark~G Ehrhart, and Britta Kilian. 2008.
\newblock A test of the factor structure equivalence of the 50-item {IPIP} five-factor model measure across gender and ethnic groups.
\newblock \emph{Journal of Personality Assessment}, 90(5):507--516.

\bibitem[{Eignor(2013)}]{eignor2013standards}
Daniel~R Eignor. 2013.
\newblock The standards for educational and psychological testing.

\bibitem[{Fitria(2023)}]{fitria2023artificial}
Tira~Nur Fitria. 2023.
\newblock Artificial intelligence ({AI}) technology in {O}pen{AI} {C}hat{GPT} application: A review of {C}hat{GPT} in writing english essay.
\newblock In \emph{ELT Forum: Journal of English Language Teaching}, volume~12, pages 44--58.

\bibitem[{Floridi and Chiriatti(2020)}]{floridi2020gpt}
Luciano Floridi and Massimo Chiriatti. 2020.
\newblock {GPT}-3: Its nature, scope, limits, and consequences.
\newblock \emph{Minds and Machines}, 30:681--694.

\bibitem[{Furr(2021)}]{Psychometrics}
R~Michael Furr. 2021.
\newblock \emph{Psychometrics: an introduction}.
\newblock SAGE publications.

\bibitem[{Gignac(2013)}]{gignac2013modeling}
Gilles~E Gignac. 2013.
\newblock Modeling the balanced inventory of desirable responding: Evidence in favor of a revised model of socially desirable responding.
\newblock \emph{Journal of Personality Assessment}, 95(6):645--656.

\bibitem[{Goldberg et~al.(2006)Goldberg, Johnson, Eber, Hogan, Ashton, Cloninger, and Gough}]{goldberg2006international}
Lewis~R Goldberg, John~A Johnson, Herbert~W Eber, Robert Hogan, Michael~C Ashton, C~Robert Cloninger, and Harrison~G Gough. 2006.
\newblock The international personality item pool and the future of public-domain personality measures.
\newblock \emph{Journal of Research in personality}, 40(1):84--96.

\bibitem[{Gow et~al.(2005)Gow, Whiteman, Pattie, and Deary}]{gow2005goldberg}
Alan~J Gow, Martha~C Whiteman, Alison Pattie, and Ian~J Deary. 2005.
\newblock Goldberg’s ‘{IPIP}’big-five factor markers: Internal consistency and concurrent validation in scotland.
\newblock \emph{Personality and Individual Differences}, 39(2):317--329.

\bibitem[{Grimm(2010)}]{sdb}
Pamela Grimm. 2010.
\newblock \href {https://doi.org/10.1002/9781444316568.wiem02057} {\emph{Social Desirability Bias}}.
\newblock John Wiley \& Sons, Ltd.

\bibitem[{Ii and Sipps(1985)}]{ii1985cross}
Avery~Zook Ii and Gary~J Sipps. 1985.
\newblock Cross-validation of a short form of the {M}arlowe-{C}rowne {S}ocial {D}esirability {S}cale.
\newblock \emph{Journal of Clinical Psychology}, 41(2):236--238.

\bibitem[{Jeong(2024)}]{jeong2024fine}
Cheonsu Jeong. 2024.
\newblock Fine-tuning and utilization methods of domain-specific {LLM}s.
\newblock \emph{arXiv preprint arXiv:2401.02981}.

\bibitem[{Kajonius(2017)}]{kajonius2017cross}
Petri~J Kajonius. 2017.
\newblock Cross-cultural personality differences between east {A}sia and northern {E}urope in {IPIP}-{NEO}.
\newblock \emph{International Journal of Personality Psychology}, 3(1):1--7.

\bibitem[{Kim et~al.(2025)Kim, Schramm, Schmitzer, Serguen, Ziegelmayer, Busch, Komenda, Makowski, Adams, Bressem et~al.}]{kim2025evaluating}
Su~Hwan Kim, Severin Schramm, Lena Schmitzer, Kerem Serguen, Sebastian Ziegelmayer, Felix Busch, Alexander Komenda, Marcus Makowski, Lisa~C Adams, Keno~K Bressem, and 1 others. 2025.
\newblock Evaluating large language model-generated brain {MRI} protocols: Performance of {GPT}-4o, o3-mini, {D}eep{S}eek-{R}1 and {Q}wen2. 5-72{B}.
\newblock \emph{medRxiv}, pages 2025--04.

\bibitem[{Kong et~al.(2023)Kong, Zhao, Chen, Li, Qin, Sun, Zhou, Wang, and Dong}]{kong2023better}
Aobo Kong, Shiwan Zhao, Hao Chen, Qicheng Li, Yong Qin, Ruiqi Sun, Xin Zhou, Enzhi Wang, and Xiaohang Dong. 2023.
\newblock Better zero-shot reasoning with role-play prompting.
\newblock \emph{arXiv preprint arXiv:2308.07702}.

\bibitem[{Lee et~al.(2024)Lee, Yang, Peng, Heo, and Liu}]{lee2024exploring}
Sanguk Lee, Kai-Qi Yang, Tai-Quan Peng, Ruth Heo, and Hui Liu. 2024.
\newblock Exploring social desirability response bias in large language models: Evidence from {GPT}-4 simulations.
\newblock \emph{arXiv preprint arXiv:2410.15442}.

\bibitem[{Li et~al.(2024{\natexlab{a}})Li, Zhang, Tang, and Li}]{li2024automatic}
Chang-Jin Li, Jiyuan Zhang, Yun Tang, and Jian Li. 2024{\natexlab{a}}.
\newblock Automatic item generation for personality situational judgment tests with large language models.
\newblock \emph{arXiv preprint arXiv:2412.12144}.

\bibitem[{Li et~al.(2024{\natexlab{b}})Li, Dong, Chen, Su, Zhou, Ai, Ye, and Liu}]{li2024llms}
Haitao Li, Qian Dong, Junjie Chen, Huixue Su, Yujia Zhou, Qingyao Ai, Ziyi Ye, and Yiqun Liu. 2024{\natexlab{b}}.
\newblock {LLM}s-as-judges: a comprehensive survey on {LLMs}-based evaluation methods.
\newblock \emph{arXiv preprint arXiv:2412.05579}.

\bibitem[{Li et~al.(2024{\natexlab{c}})Li, Zhang, Li, Sun, and Brown}]{li2024mixed}
Mengtong Li, Bo~Zhang, Lingyue Li, Tianjun Sun, and Anna Brown. 2024{\natexlab{c}}.
\newblock Mixed-keying or desirability-matching in the construction of forced-choice measures? an empirical investigation and practical recommendations.
\newblock \emph{Organizational Research Methods}, page 10944281241229784.

\bibitem[{Madaan et~al.(2023)Madaan, Tandon, Gupta, Hallinan, Gao, Wiegreffe, Alon, Dziri, Prabhumoye, Yang et~al.}]{madaan2023self}
Aman Madaan, Niket Tandon, Prakhar Gupta, Skyler Hallinan, Luyu Gao, Sarah Wiegreffe, Uri Alon, Nouha Dziri, Shrimai Prabhumoye, Yiming Yang, and 1 others. 2023.
\newblock Self-refine: Iterative refinement with self-feedback.
\newblock \emph{Advances in Neural Information Processing Systems}, 36:46534--46594.

\bibitem[{McCrae and Costa(1987)}]{mccrae1987validation}
Robert~R McCrae and Paul~T Costa. 1987.
\newblock Validation of the five-factor model of personality across instruments and observers.
\newblock \emph{Journal of personality and social psychology}, 52(1):81.

\bibitem[{McCrae et~al.(2005)McCrae, Costa, and Martin}]{mccrae2005neo}
Robert~R McCrae, Paul~T Costa, Jr, and Thomas~A Martin. 2005.
\newblock The {NEO--PI--}3: A more readable revised {NEO} personality inventory.
\newblock \emph{Journal of personality assessment}, 84(3):261--270.

\bibitem[{{OpenAI}(2025)}]{openai2025o3system}
{OpenAI}. 2025.
\newblock \href {https://cdn.openai.com/pdf/2221c875-02dc-4789-800b-e7758f3722c1/o3-and-o4-mini-system-card.pdf} {{O}pen{AI} o3 and o4-mini system card}.
\newblock System card, OpenAI, San Francisco, CA.
\newblock Version 2 of the Preparedness Framework.

\bibitem[{Park et~al.(2024)Park, Schoenegger, and Zhu}]{park2024diminished}
Peter~S Park, Philipp Schoenegger, and Chongyang Zhu. 2024.
\newblock Diminished diversity-of-thought in a standard large language model.
\newblock \emph{Behavior Research Methods}, 56(6):5754--5770.

\bibitem[{Patel et~al.(2023)Patel, Raut, Zimlichman, Cheetirala, Nadkarni, Glicksberg, Freeman, Timsina, and Klang}]{patel2023limits}
Dhavalkumar Patel, Ganesh Raut, Eyal Zimlichman, Satya~Narayan Cheetirala, Girish Nadkarni, Benjamin~S Glicksberg, Robert Freeman, Prem Timsina, and Eyal Klang. 2023.
\newblock The limits of prompt engineering in medical problem-solving: a comparative analysis with {C}hat{GPT} on calculation based {USMLE} medical questions.
\newblock \emph{MedRxiv}, pages 2023--08.

\bibitem[{Paulhus and Reid(1991)}]{paulhus1991enhancement}
Delroy~L Paulhus and Douglas~B Reid. 1991.
\newblock Enhancement and denial in socially desirable responding.
\newblock \emph{Journal of personality and social psychology}, 60(2):307.

\bibitem[{Podsakoff et~al.(2003)Podsakoff, MacKenzie, Lee, and Podsakoff}]{podsakoff2003common}
Philip~M Podsakoff, Scott~B MacKenzie, Jeong-Yeon Lee, and Nathan~P Podsakoff. 2003.
\newblock Common method biases in behavioral research: a critical review of the literature and recommended remedies.
\newblock \emph{Journal of applied psychology}, 88(5):879.

\bibitem[{Putnick and Bornstein(2016)}]{putnick2016measurement}
Diane~L Putnick and Marc~H Bornstein. 2016.
\newblock Measurement invariance conventions and reporting: The state of the art and future directions for psychological research.
\newblock \emph{Developmental review}, 41:71--90.

\bibitem[{Reynolds(1982)}]{reynolds1982development}
William~M Reynolds. 1982.
\newblock Development of reliable and valid short forms of the {M}arlowe-{C}rowne social desirability scale.
\newblock \emph{Journal of clinical psychology}, 38(1):119--125.

\bibitem[{Ryoo et~al.(2022)Ryoo, Park, Suh, Choi, and Kwon}]{ryoo2022development}
Ji~Hoon Ryoo, Sunhee Park, Hongwook Suh, Jaehwa Choi, and Jongkyum Kwon. 2022.
\newblock Development of a new measure of cognitive ability using automatic item generation and its psychometric properties.
\newblock \emph{Sage Open}, 12(2):21582440221095016.

\bibitem[{Sahoo et~al.(2024)Sahoo, Singh, Saha, Jain, Mondal, and Chadha}]{sahoo2024systematic}
Pranab Sahoo, Ayush~Kumar Singh, Sriparna Saha, Vinija Jain, Samrat Mondal, and Aman Chadha. 2024.
\newblock A systematic survey of prompt engineering in large language models: Techniques and applications.
\newblock \emph{arXiv preprint arXiv:2402.07927}.

\bibitem[{Salecha et~al.(2024)Salecha, Ireland, Subrahmanya, Sedoc, Ungar, and Eichstaedt}]{salecha2024large}
Aadesh Salecha, Molly~E Ireland, Shashanka Subrahmanya, Jo{\~a}o Sedoc, Lyle~H Ungar, and Johannes~C Eichstaedt. 2024.
\newblock Large language models show human-like social desirability biases in survey responses.
\newblock \emph{arXiv preprint arXiv:2405.06058}.

\bibitem[{Satterfield et~al.(2024)Satterfield, Holbrooka, and Wilcoxa}]{satterfield2024fine}
Nolan Satterfield, Parker Holbrooka, and Thomas Wilcoxa. 2024.
\newblock Fine-tuning llama with case law data to improve legal domain performance.
\newblock \emph{OSF Preprints}.

\bibitem[{Schneider(2024)}]{schneider2024explainable}
Johannes Schneider. 2024.
\newblock Explainable generative {AI} ({G}en{XAI}): A survey, conceptualization, and research agenda.
\newblock \emph{Artificial Intelligence Review}, 57(11):289.

\bibitem[{Shin et~al.(2025)Shin, Lee, and Kim}]{shin2025exploratory}
Dongkwang Shin, Jang~Ho Lee, and Kyungmin Kim. 2025.
\newblock An exploratory study on two automated item generators for generating {L}2 reading test items.
\newblock \emph{RELC Journal}, page 00336882251326284.

\bibitem[{Tao et~al.(2009)Tao, Guoying, and Brody}]{tao2009preliminary}
Peng Tao, Dong Guoying, and Stuart Brody. 2009.
\newblock Preliminary study of a {C}hinese language short form of the marlowe--crowne social desirability scale.
\newblock \emph{Psychological reports}, 105(3\_suppl):1039--1046.

\bibitem[{Ullah et~al.(2024)Ullah, Parwani, Baig, and Singh}]{ullah2024challenges}
Ehsan Ullah, Anil Parwani, Mirza~Mansoor Baig, and Rajendra Singh. 2024.
\newblock Challenges and barriers of using large language models ({LLM}) such as chatgpt for diagnostic medicine with a focus on digital pathology--a recent scoping review.
\newblock \emph{Diagnostic pathology}, 19(1):43.

\bibitem[{Wei et~al.(2022)Wei, Wang, Schuurmans, Bosma, Xia, Chi, Le, Zhou et~al.}]{wei2022chain}
Jason Wei, Xuezhi Wang, Dale Schuurmans, Maarten Bosma, Fei Xia, Ed~Chi, Quoc~V Le, Denny Zhou, and 1 others. 2022.
\newblock Chain-of-thought prompting elicits reasoning in large language models.
\newblock \emph{Advances in neural information processing systems}, 35:24824--24837.

\bibitem[{Xue et~al.(2025{\natexlab{a}})Xue, Huang, Ji, and Wang}]{iteration2}
Eric Xue, Zeyi Huang, Yuyang Ji, and Haohan Wang. 2025{\natexlab{a}}.
\newblock Improve: Iterative model pipeline refinement and optimization leveraging llm agents.
\newblock \emph{arXiv preprint arXiv:2502.18530}.

\bibitem[{Xue et~al.(2025{\natexlab{b}})Xue, Liu, and Xiong}]{pmlr-v264-xue25a}
Mingfeng Xue, Yunting Liu, and HuaXia Xiong. 2025{\natexlab{b}}.
\newblock \href {https://proceedings.mlr.press/v264/xue25a.html} {Enhancing non-cognitive assessments with {GPT}: Innovations in item generation and translation for the university belonging questionnaire}.
\newblock In \emph{Proceedings of Large Foundation Models for Educational Assessment}, volume 264 of \emph{Proceedings of Machine Learning Research}, pages 157--172. PMLR.

\bibitem[{Yang et~al.(2025)Yang, Kleinman, and Harteveld}]{gpt4game}
Daijin Yang, Erica Kleinman, and Casper Harteveld. 2025.
\newblock \href {https://doi.org/10.1109/TG.2025.3563780} {{GPT} for games: An updated scoping review (2020-2024)}.
\newblock \emph{IEEE Transactions on Games}, pages 1--16.

\bibitem[{Ypofanti et~al.(2015)Ypofanti, Zisi, Zourbanos, Mouchtouri, Tzanne, Theodorakis, and Lyrakos}]{ypofanti2015psychometric}
Maria Ypofanti, Vasiliki Zisi, Nikolaos Zourbanos, Barbara Mouchtouri, Pothiti Tzanne, Yannis Theodorakis, and Georgios Lyrakos. 2015.
\newblock Psychometric properties of the international personality item pool big-five personality questionnaire for the greek population.
\newblock \emph{Health psychology research}, 3(2):2206.

\bibitem[{Zhang et~al.(2025)Zhang, Huang, Sun, and Savalei}]{zhang2025improving}
Xijuan Zhang, Muhua Huang, Jessie Sun, and Victoria Savalei. 2025.
\newblock Improving the measurement of the big five via alternative formats for the bfi-2.
\newblock \emph{Journal of Personality Assessment}, pages 1--22.

\bibitem[{Zheng et~al.(2008)Zheng, Goldberg, Zheng, Zhao, Tang, and Liu}]{zheng2008reliability}
Lijun Zheng, Lewis~R Goldberg, Yong Zheng, Yufang Zhao, Yonglong Tang, and Li~Liu. 2008.
\newblock Reliability and concurrent validation of the {IPIP} big-five factor markers in {C}hina: Consistencies in factor structure between internet-obtained heterosexual and homosexual samples.
\newblock \emph{Personality and individual differences}, 45(7):649--654.

\end{thebibliography}
\end{document}